\documentclass[letterpaper, 10 pt, conference]{ieeeconf}  
                                                          
\IEEEoverridecommandlockouts                              
                                               
\usepackage{graphicx}
\usepackage{amsmath}
\usepackage{bm}
\usepackage{url}
\usepackage{subcaption}

\title{\LARGE \bf
Mixed Reality as Communication Medium for Human-Robot Collaboration
}

\author{Simone~Macci\`o, Alessandro~Carf\`i, and Fulvio~Mastrogiovanni
\thanks{This work has been supported by the European Union Erasmus+ Programme via the Joint Master Degree program European Master on Advanced Robotics Plus (EMARO+), and via the European Union CHIST-ERA program (2014-2020) project InDex (Robot In-hand Dexterous Manipulation).}
\thanks{All the authors are with the Department of Informatics, Bioengineering, Robotics, and Systems Engineering, University of Genoa, Via Opera Pia 13, 16145 Genoa, Italy (emails: simone.maccio@edu.unige.it; alessandro.carfi@dibris.unige.it; fulvio.mastrogiovanni@unige.it)}%
}

\begin{document}

\onecolumn
© 2022 IEEE.  Personal use of this material is permitted.  Permission from IEEE must be obtained for all other uses, in any current or future media, including reprinting/republishing this material for advertising or promotional purposes, creating new collective works, for resale or redistribution to servers or lists, or reuse of any copyrighted component of this work in other works.
\newpage
\twocolumn
\maketitle
\pagestyle{empty}

\begin{abstract}

Humans engaged in collaborative activities are naturally able to convey their intentions to teammates through multi-modal communication, which is made up of explicit and implicit cues. 
Similarly, a more natural form of human-robot collaboration may be achieved by enabling robots to convey their \textit{intentions} to human teammates via multiple communication channels. 
In this paper, we postulate that a better communication may take place should collaborative robots be able to anticipate their movements to human teammates in an intuitive way.
In order to support such a claim, we propose a robot system's architecture through which robots can communicate planned motions to human teammates leveraging a Mixed Reality interface powered by modern head-mounted displays. 
Specifically, the robot's hologram, which is superimposed to the real robot in the human teammate's point of view, shows the robot's \textit{future} movements, allowing the human to understand them in advance, and possibly react to them in an appropriate way. 
We conduct a preliminary user study to evaluate the effectiveness of the proposed anticipatory visualization during a complex collaborative task. 
The experimental results suggest that an improved and more natural collaboration can be achieved by employing this \textit{anticipatory communication mode}.

\end{abstract}


\section{Introduction}
\label{section1}

In the past few years, the guidelines of Industry 4.0, while envisioning a shift towards a human-centric paradigm, demanded the development of a new generation of robots capable of working alongside humans. 
Technological advancements in multi-modal perception, reasoning, control, and actuation \cite{ruffaldi2016third} endowed such modern robot platforms with higher degrees of safety, environmental awareness, and flexibility, therefore allowing them to coexist and collaborate (hence the name \textit{cobots}) with human teammates inside a shared and unstructured workspace \cite{ajoudani2018progress}. 

\begin{figure}[t!]
\centering
\includegraphics[width=0.48\textwidth]{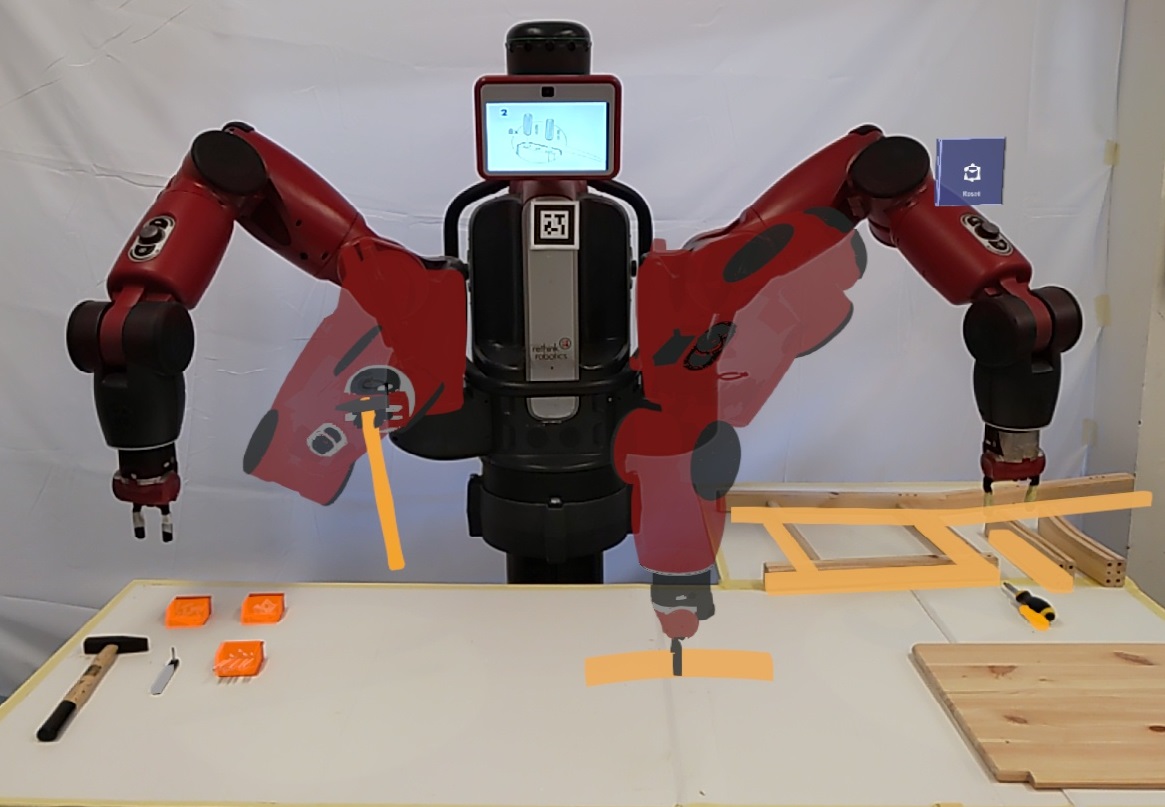}
\caption{
First person view of the Mixed Reality interface seen through an head-mounted display.
The holographic representation allows the human to see upcoming robot motions during a collaborative process.}
\label{fig::mrinterfacefigure}
\end{figure}

In order to achieve an adequate level of collaboration, effective communication is of the utmost importance. 
Communication involving humans is intrinsically multi-modal, and results from the interplay of explicit and implicit communication channels, or \textit{media}. 
In a human-human collaboration process, involved individuals are typically able to infer each other's intentions through differing channels, which involve explicit signals, e.g., speech, and implicit cues, e.g., gaze, posture and gestures \cite{klein2005common, mutlu2009nonverbal, calisgan2012identifying}, just to name a few.

In this paper, we refer generically to the notion of \textit{piece of information} to designate the meaningful content of such explicit or implicit cues, conveyed via one or more \textit{communicative acts}. 
A naïve use of these terms should suffice for the discussion hereinafter.  
Given a piece of information $I$, we provide an operative formalization of the associated communicative act $C$ as
\begin{equation}
C = f_{m_i}(I, \bm{t}_{m_i}),
\label{eq:comm_act_1}
\end{equation}
where $m_i \in M$ is the medium selected from the set of all possible communicative media $M = \{m_0, \dots, m_{|M|}\}$, $f_{m_i}$ is a medium-specific function \textit{rendering} the information into a communicative act, and $\bm{t} = \{t_{m_i,s}, \dots, t_{m_i,e}; \leq\}$ is an ordered set delimiting the start and end time instants of the communicative act. 
In a more general sense, communicative acts can result from the interplay of $N \leq |M|$ different media, and therefore we can reformulate \eqref{eq:comm_act_1} as
\begin{equation}
C = \bigcup_{i=1}^{N \leq |M|} f_{m_i}(I, \bm{t}_{m_i}). 
\label{comm_act}
\end{equation}

It is noteworthy that not all communicative media have the same effectiveness for a given interaction process, and the combination of multiple media is expected to increase clarity and reduce ambiguity.
However, it is not the aim of this paper to discuss in depth such possibilities, whereas we will limit our focus to a simple case.
In order to equip collaborative robots with an expanded range of communicative media, the literature explored different solutions, including screens mounted in working environments \cite{shrestha2016intent}, light signals associated with the next robot action \cite{cha2016nonverbal}, or natural language interfaces \cite{gong2018behavior}.
Nevertheless, these methods suffer from several environmental and contextual constraints: 
external screens force human teammates to continuously divert their attention from the shared workspace, light cues may result counter-intuitive, and verbal cues are not appropriate in noisy industrial scenarios.

A promising alternative is represented by Mixed Reality (MR) \cite{milgram1994taxonomy}, i.e., an hybrid communicative medium resulting from the combination of real physical environment and virtual holographic objects (Figure \ref{fig::mrinterfacefigure}). Unlike Augmented Reality (AR), which is limited to the overlay of digital visual content onto the observed scene, MR creates immersive experiences by bringing interactive and \textit{spatially contextualized}  holograms into the real world.
Traditionally, MR interfaces in Human-Robot Collaboration (HRC) are achieved in combination with handheld tablets, as in \cite{frank2016realizing}, to show 3D holograms superimposed to the camera's feed, or with projectors mounted above the shared workspace to display 2D safety boundaries for the human teammate \cite{vogel2017safeguarding}. 
However, the former approach intrinsically limits the worker freedom to use hands during the collaboration process, whereas the latter requires a precisely structured environment, a complex calibration, and only provides the user with 2D information.

The introduction of commercial MR Head Mounted Displays (MR-HMDs), such as the Microsoft \textit{HoloLens}
family, enabled researchers to exploit the MR medium in HRC scenarios, providing human teammates with the possibility of perceiving the holographic representation from a first-person perspective, while keeping their hands free \cite{quintero2018robot, wang2020closed}. 
Some studies suggested to use MR to display a static representation of robot's future movements to anticipate whether a particular trajectory is collision free \cite{rosen2019communicating, gruenefeld2020mind}, or to perform an handover \cite{newbury2021visualizing}. 
These approaches show how using MR to represent future robot motions improves the overall interaction efficiency and fluency. 

The effect of anticipatory motions can be framed as a combination of communicative acts in the form of \eqref{comm_act}.
First, we formalize the contribute of robot motions as communicative acts. 
We observe that a collaborative robot already communicates pieces of information through its own motion, formally represented as a trajectory $\bm{\tau}$ in the robot joint space $\bm{q}$, and spanning $\bm{t}_{m}$ time instants.
Then, starting from the general relation in \eqref{comm_act}, considering the robot motion as a possible communicative medium \textit{m}, setting $N=1$, and posing $m_i = m$, we derive its instance $\hat{C}$ as
\begin{equation}
\hat{C} = f_{m}(I, \bm{t}_{m}) = \bm{\tau}(\bm{t}_{m}),
\label{eq:comm_act_m}
\end{equation}
where
\begin{equation}
\bm{\tau}(\bm{t}_{m}) = \{q_{0}(\bm{t}_{m}), \ldots, q_{|q|}(\bm{t}_{m})\}.
\end{equation}
Then, we model the combination of robot and anticipatory motions via MR.
This is done by joining two communicative acts, which result from two media, i.e., robot motion \textit{m} and the \textit{anticipated} robot motion \textit{am}, which can be provided by an holographic representation in MR, as
\begin{equation}
\hat{C}^* = \bm{\tau}(\bm{t}_{m}) \cup \bm{\tau}(\bm{t}_{am}),
\label{comm_holo}
\end{equation}
with
\begin{equation}
\bm{t}_{am} = \bm{t}_{m} - \Delta t, 
\label{comm_holo_time}
\end{equation}
where $\Delta t$ is the time interval from the initial time instant of the holographic, anticipated robot trajectory in \textit{am} and the related robot motion in \textit{m}.

In this paper, we argue that the interplay between these two media can improve team efficiency and the fluency of the collaboration by reducing communication ambiguities. 
Differently from the literature discussed above, where human-robot collaboration is limited to relatively simple, repetitive tasks, with little to no teamwork involved, in this paper we aim at studying the effectiveness of the whole communicative act $\hat{C}^*$ in a \textit{structured} HRC scenario. 
Our contribution is two-fold: 
i) a publicly available, modular software architecture, which we call MR-HRC, allowing human teammates to preview robot intended motions via MR as envisioned by $\hat{C}^*$; 
ii) a user study carried out within a collaborative assembly scenario devised to compare the effectiveness of $\hat{C}$ in \eqref{eq:comm_act_m} and $\hat{C}^*$ in \eqref{comm_holo}. 
To this extent, we hypothesize that the communication act $\hat{C}^*$, resulting from the interplay between the robot motion and the anticipatory robot motion media, is more effective than $\hat{C}$ alone. 
In particular, we hypothesize that $\hat{C}^*$:
\begin{itemize}
\item[\textit{H1}] reduces the number of accidental collisions during the collaboration process and, consequently, (hypothesis \textit{H1.a}) collaboration downtime is reduced as well;
\item[\textit{H2}] increases the human teammate's proactivity. 
\end{itemize}

The rest of the paper is organized as follows. 
Section \ref{sec:softarchitecture} describes the general structure of MR-HRC implementing $\hat{C}^*$.
Section \ref{sec:implementation} provides the implementation details of MR-HRC, along with the used hardware setup.
Section \ref{sec:experimentalsetup} describes the experimental scenario, whereas Section \ref{sec:results} discusses the results obtained in a preliminary user study.
Finally, Section \ref{sec:conclusions} illustrates the conclusions.


\section{System's Architecture}
\label{sec:softarchitecture}

\begin{figure*}[t!]
\centering \vspace{0.15cm}
\includegraphics[width=0.95\textwidth]{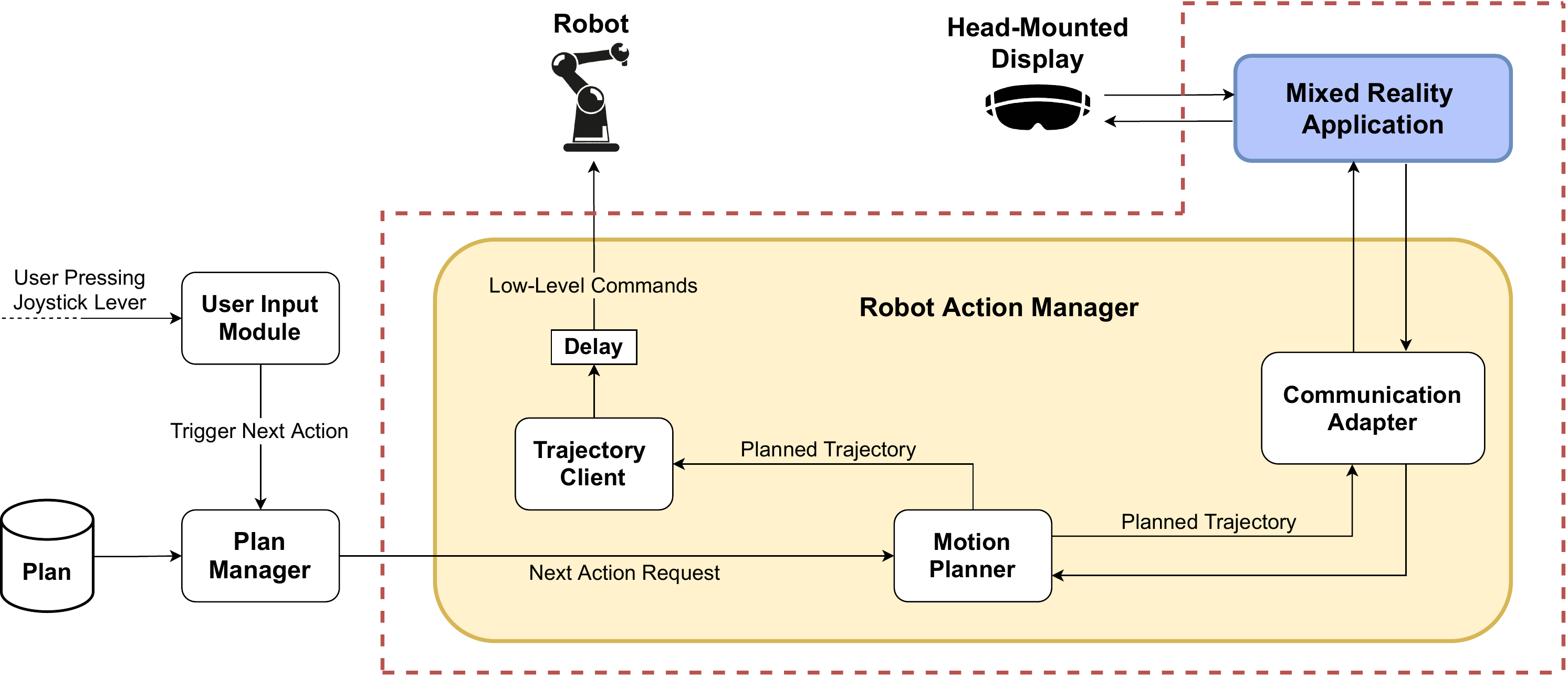}
\caption{
The \textit{Mixed Reality Application} and the \textit{Robot Action Manager} are the core components of MR-HRC. 
\textit{Motion Planner} computes robot movements and dispatches planned trajectories to both \textit{Mixed Reality Application} for the \textit{am} medium, and \textit{Trajectory Client}, which deals with actual motion execution on the real robot in the \textit{m} medium after introducing a delay $\Delta t$. 
\textit{Plan Manager} decides the next robot action and, as such, acts as an external high-level task planner.}
\label{fig:systemarchitecturefigure}
\end{figure*}

This Section describes the core of MR-HRC, which is depicted on the right-hand side of Figure \ref{fig:systemarchitecturefigure}, enclosed within dashed lines.
The left-hand side of the Figure shows how this core has been integrated in an expanded architecture used in the experimental scenario, as shown in Section \ref{sec:experimentalsetup}.
Implementation details are discussed in Section \ref{sec:implementation}. 

Two are the core components of MR-HRC.
On the one hand, a module named \textit{Mixed Reality Application} deployed on the HMD renders the MR interface enabling the human teammate to perceive the holograms superimposed to real world objects. 
Conversely, the high-level component \textit{Robot Action Manager} groups all the modules dealing with robot motion planning and execution. 
The two components are interfaced through an appropriate \textit{Communication Adapter}.


\textit{Mixed Reality Application} acts as virtual counterpart of the human-robot collaborative scenario. 
As depicted in Figure \ref{fig::mrinterfacefigure}, this module aims at displaying through the HMD the robot holographic representation, as well as the holograms of relevant tools and objects involved in the collaborative process. 
Focusing on the robot hologram, whenever a new robot action is planned, a contrived delay $\Delta t$ is introduced as foreseen in \eqref{comm_holo} and \eqref{comm_holo_time} to allow the robot hologram to anticipate the robot motions as provisioned by the \textit{am} medium,
thus enabling human teammates to experience the holographic robot action in advance.


In order to obtain a consistent MR experience, the first step involves calibrating the two media \textit{m} and \textit{am}, i.e., the real and the holographic robot's reference frames. 
For this purpose, we employ a simple 2D bar-code marker attached to the robot, and exploit one of the many available software modules providing marker detection and 3D pose estimation capabilities. 
Upon detecting the marker, the module returns its 3D coordinates with respect to the HMD's frontal camera and such position is used to spawn and anchor the holograms coherently with the actual scene. 
We assume that the robot's base is fixed throughout the whole collaboration process, and therefore a continuous tracking of the marker is not necessary.
The 2D marker employed in our experimental scenario is visible in Figure \ref{fig::mrinterfacefigure}, attached to the robot's front. 



\textit{Robot Action Manager} is a collection of modules collaboratively enabling robot action planning and execution.
The functionalities of such modules are detailed hereby, along with brief explanations on how they interact with one another and with the \textit{Mixed Reality Application}.

\textit{Communication Adapter} establishes a serial connection between \textit{Mixed Reality Application} and \textit{Motion Planner} for data exchange. 
The module handles all the communication between these two endpoints, including service requests or messages broadcasting, it manages data serialization and deserialization, and forwards all data through the network. 

\begin{figure*}[t!]
\centering \vspace{0.15cm}
\begin{subfigure}{.325\textwidth}
  \centering
  \includegraphics[width=\linewidth]{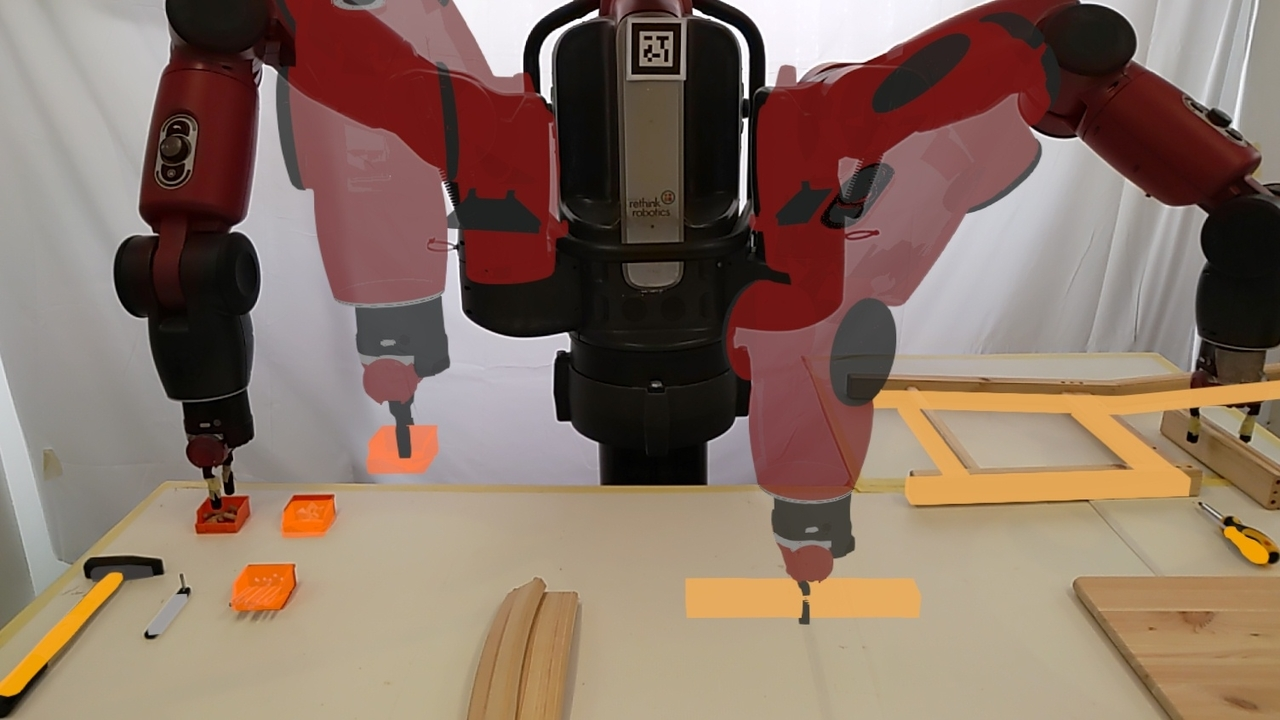}  
  \caption{}
  \label{holo1}
\end{subfigure}
\begin{subfigure}{.325\textwidth}
  \centering
  \includegraphics[width=\linewidth]{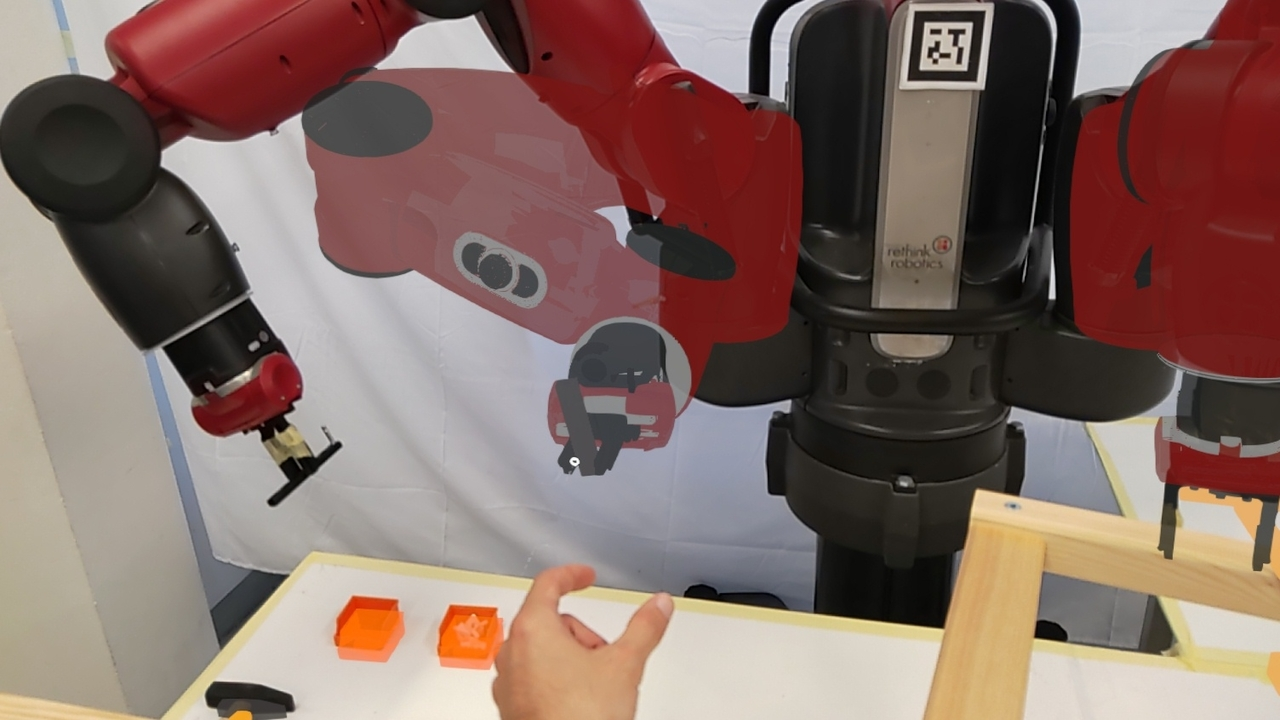}  
  \caption{}
  \label{holo2}
\end{subfigure}
\begin{subfigure}{.325\textwidth}
  \centering
  \includegraphics[width=\linewidth]{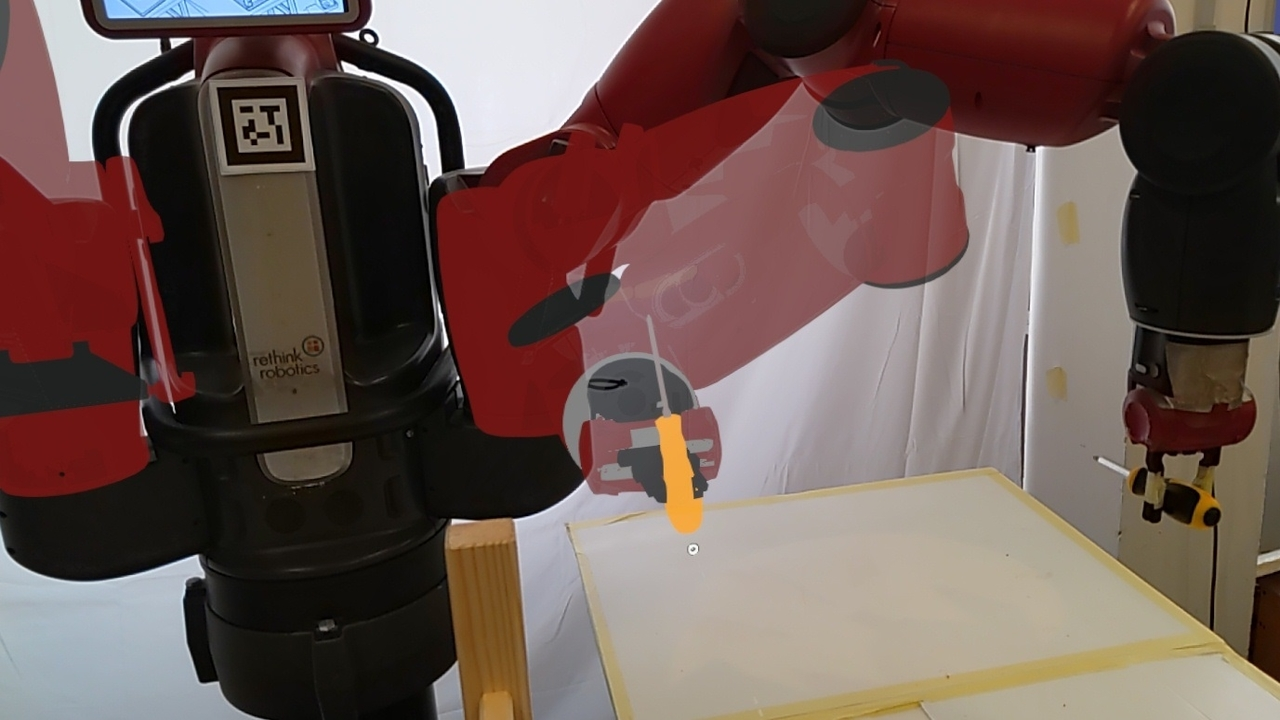}  
  \caption{}
  \label{holo3}
\end{subfigure}
\begin{subfigure}{.325\textwidth}
  \centering
  \includegraphics[width=\linewidth]{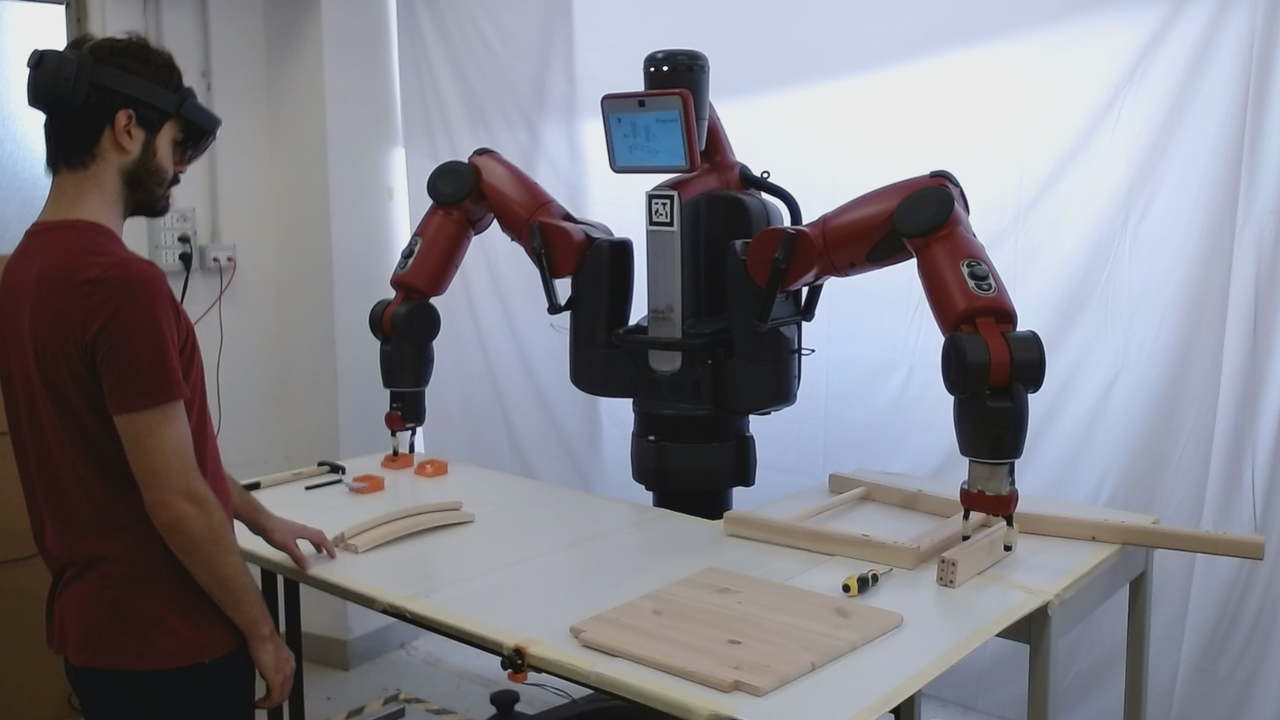}  
  \caption{}
  \label{ext1}
\end{subfigure}
\begin{subfigure}{.325\textwidth}
  \centering
  \includegraphics[width=\linewidth]{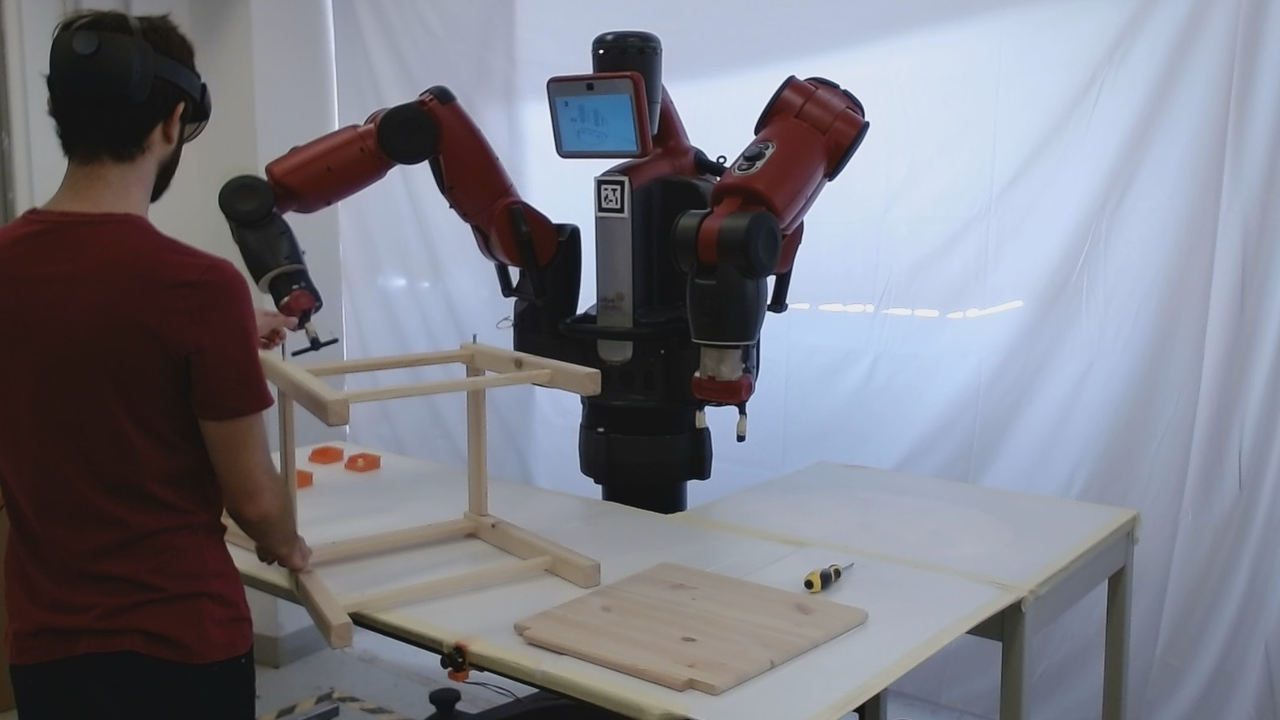}  
  \caption{}
  \label{ext2}
\end{subfigure}
\begin{subfigure}{.325\textwidth}
  \centering
  \includegraphics[width=\linewidth]{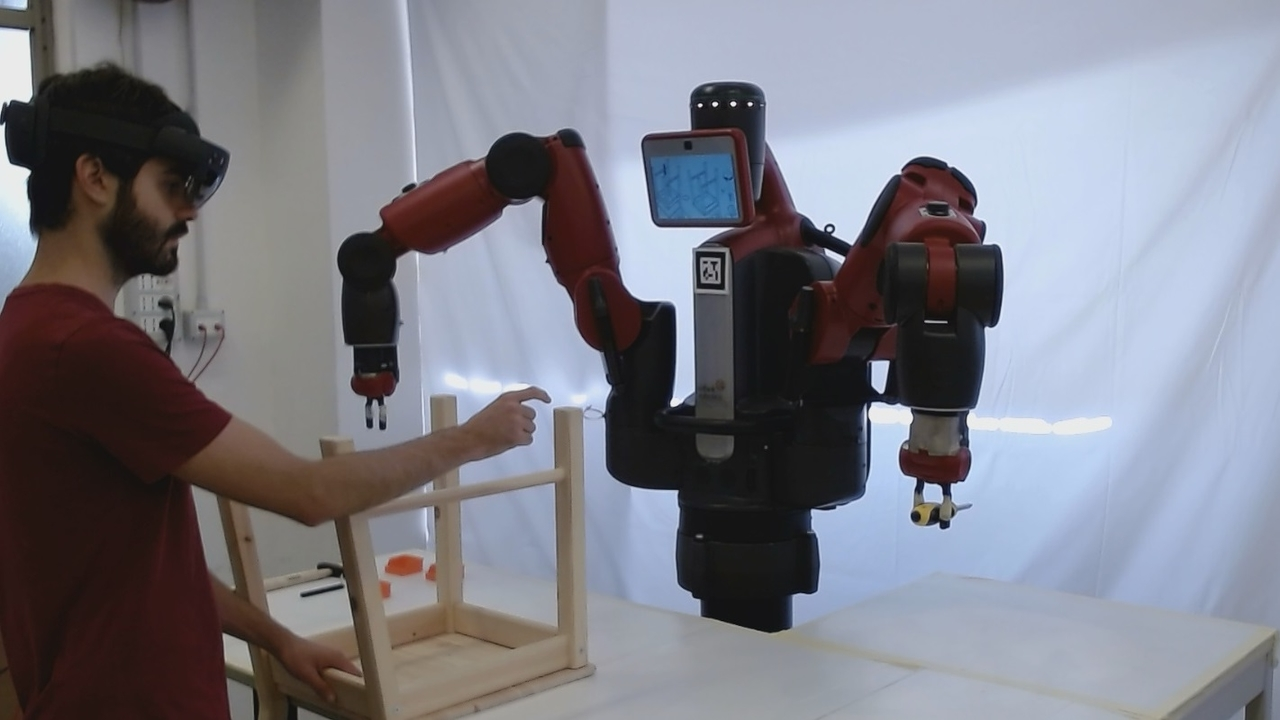}  
  \caption{}
  \label{ext3}
\end{subfigure}
\caption{A collaborative assembly process from two perspectives: top images show the first person view with the anticipatory holograms, whereas bottom images depict the same instants from an external point of view. In particular, in (a) and (d) Baxter is delivering wooden pieces and a box of dowels, in (b) and (e) Baxter supplies the hex key, and in (c) and (f) Baxter is handing over a screwdriver.}
\label{fig:exp_screenshots}
\end{figure*}

\textit{Motion Planner} is responsible for planning robot motions. 
With the aim of broadening the spectrum of possible applications for MR-HRC, we assume the robot to be capable of performing different types of action (i.e., \textit{skills}), each with its own specific motion planning routine. 
In particular, due to the collaborative scenario we have considered for this work, which involves an assembly, we focused on \textit{pick-and-place} and \textit{handover} actions, both types of action requiring the robot to fetch objects, but differing in how such items are supplied to the human teammate. 
In the case of pick-and-place actions, the robot retrieves objects needed for the collaboration process, and places them in a predefined position within the workspace, as in Figure \ref{holo1}, whereas through handovers it delivers tools in such a way that the human teammate can comfortably grasp them, as shown in Figure \ref{holo2} or Figure \ref{holo3}. 
For this reason, a planning request made to \textit{Motion Planner} by an external high-level task planner, such as \textit{Plan Manager}, must specify two parameters, namely the type of action for which a plan is sought, along with an identifier (ID) representing the requested object or tool to be fetched. 
Upon receiving a planning request, \textit{Motion Planner} communicates with \textit{Mixed Reality Application} to retrieve the object's pose (i.e., position and orientation) based on its ID, and then it uses such pose to plan the robot's motion for that particular action. 
Once the motion plan is computed, it is sent back in the form of an array of joint configurations which make up the robot's trajectory $\bm{\tau}(\bm{t}_{am})$ to \textit{Mixed Reality Application}, i.e., the robot communicative act, where it is processed and rendered as holographic motion of the robot's arm in the \textit{am} medium.
At the same time, the planned trajectory $\bm{\tau}(\bm{t}_{m})$ is forwarded to \textit{Trajectory Client}, which instead deals with execution of the motion on the real robot, i.e., in the \textit{m} medium.
As described above, this coordinated execution of trajectories in the robot motion medium and in the anticipated robot motion medium constitute the combined communicative act in \eqref{comm_holo}.

Finally, \textit{Trajectory Client} is the module performing the translation from the \textit{Motion Planner}'s output to a series of low-level commands for the joint-level controller, thus enabling the execution of the planned trajectory $\bm{\tau}(\bm{t}_{m})$ on the real robot. 
The commands are delayed for $\Delta t$ seconds (which is a parameter set to $3$ seconds in actual experiments), so that the human teammate can visualize the holographic action first. 
After such interval, the robot's arm begins moving following its virtual counterpart.


\section{Implementation, Frameworks and Equipment}
\label{sec:implementation}



The \textit{Mixed Reality Application} component has been developed using Unity, a worldwide popular game engine, which is supported by most commercial HMDs. 
The marker detection pipeline is managed by \textit{Vuforia}, a well-known software development kit used to create MR contents easily integrated with Unity. 
The holographic representation is instead built using the Microsoft \textit{Mixed Reality Toolkit}\footnote{Web: {https://github.com/microsoft/MixedRealityToolkit-Unity}.} (MRTK), a collection of tools and libraries specifically developed to design applications for MR-HMDs in Unity. 
MRTK yields the instruments to transform a standard 2D Unity scene into a 3D MR and provides the building blocks necessary to design augmented user interfaces. 
In this work, MRTK has been used to create the 3D MR interface by overlaying the virtual models onto the real objects, according to the coordinates returned by the Vuforia's marker detection pipeline.
The robot model is incorporated into the Unity scene through the Unified Robot Description Format (URDF) Importer package\footnote{Web: {https://github.com/Unity-Technologies/URDF-Importer}.}, developed by Unity Technologies, and then rendered as a 3D MR asset through MRTK.

The Robot Operating System (ROS) \cite{quigley2009ros} has been employed to develop the whole \textit{Robot Action Manager} stack. 
In particular, we used the \textit{MoveIt} \cite{chitta2012moveit} framework to implement the \textit{Motion Planner} module. 
The \textit{Communication Adapter}, instead, has been developed exploiting the ROS-Unity integration package\footnote{Web: {https://github.com/Unity-Technologies/ROS-TCP-Endpoint}.}, which has been recently published by Unity Technologies. 

In order to support other researchers interested in the topic, we have decided to make the code of our architecture publicly available on GitHub\footnote{Web: {https://github.com/TheEngineRoom-UniGe/MixedRealityHRC.git}.}.


As far as the equipment is concerned, a Microsoft \textit{HoloLens 2} headset has been used to run the holographic medium. 
This powerful HMD natively supports MR applications developed through Unity and possesses a wide range of sensors useful for spatial perception, including an array of four cameras used for head tracking and a time-of-flight (ToF) camera for depth sensing. 
Finally, the device's screen provides a $52^{\circ}$ diagonal field of view for holograms projection, thus making it suitable to display digital overlays even in case of a close-proximity collaboration process with a robot.

The employed robot platform is Baxter from Rethink Robotics \cite{fitzgerald2013developing}, a well-known dual-arm manipulator, which has been previously employed in related research works \cite{ruffaldi2016third, rosen2019communicating}. 
In order to ensure that communication could be established between the ROS environment running within the Baxter's embedded computer and the Unity app deployed on HoloLens, the robot and the HMD have been connected to the same local network. 


\section{Experimental Setup} 
\label{sec:experimentalsetup}


\subsection{Collaborative Scenario} 
\label{collaborativescenario}

The target task considered in this paper consists in the collaborative assembly of a wooden chair.
The workspace whereby the human-robot collaboration process takes place is visible in Figure \ref{fig:exp_screenshots}. 
It includes a table which use is shared by both the human and the robot teammates, on top of which  various wooden pieces, components, and such tools as a screwdriver, a hammer, and an hex key are placed.
The assembly components, such as screws, dowels and bolts, are stored in orange boxes, which the robot can supply with during the collaborative process. 

The task is divided in $10$ sequential steps.
During each step Baxter performs pick-and-place or handover operations, providing its human teammate with the items necessary to build the piece of furniture step by step, and putting tools and components away when they are not needed anymore. 
The assembly actions, instead, are strictly left to the human teammate. 
In order to provide the human teammate  with support during the assembly phase, step by step visual instructions are shown on the Baxter's \textit{head} display.
It is worth mentioning that the typical trial lasts around $10$ minutes, during which the human and the robot teammates keep collaborating on the assembly\footnote{An exemplar run of the collaboration process is available at the following: https://youtu.be/uXiH9ElsiD4}. 

In order to adapt the system's architecture to this peculiar scenario, two additional software components have been implemented and added, i.e., \textit{Plan Manager} and \textit{User Input Module}.
\textit{Plan Manager} handles the overall Baxter's plan, which for the present purposes consists in a sequence of scripted actions that the robot carries out during the collaborative task. 
The plan is stored as a text file and, due to the sequential nature of the assembly process, the actions are performed one by one in a fixed order. 
The module waits for an input from the human teammate, then reads the next action to carry out from the sequence, which specifies what the robot has to do next, along with the ID of the required object to fetch, and finally sends a request to \textit{Motion Planner} with such parameters.
On the other hand, the \textit{User Input Module} listens to human inputs through an analog joystick module mounted on the human worker's side of the workspace. 
Whenever human teammates have completed an assembly step and are idle, they can press the joystick lever, thus sending a Boolean command to \textit{Plan Manager}, which in turn publishes the next action message.



\begin{figure*}[t!]
\centering \vspace{0.15cm}
\begin{subfigure}{.317\textwidth}
  \centering
  \includegraphics[width=\linewidth, trim={0.7cm 0.8cm 1.7cm 1.5cm},clip]{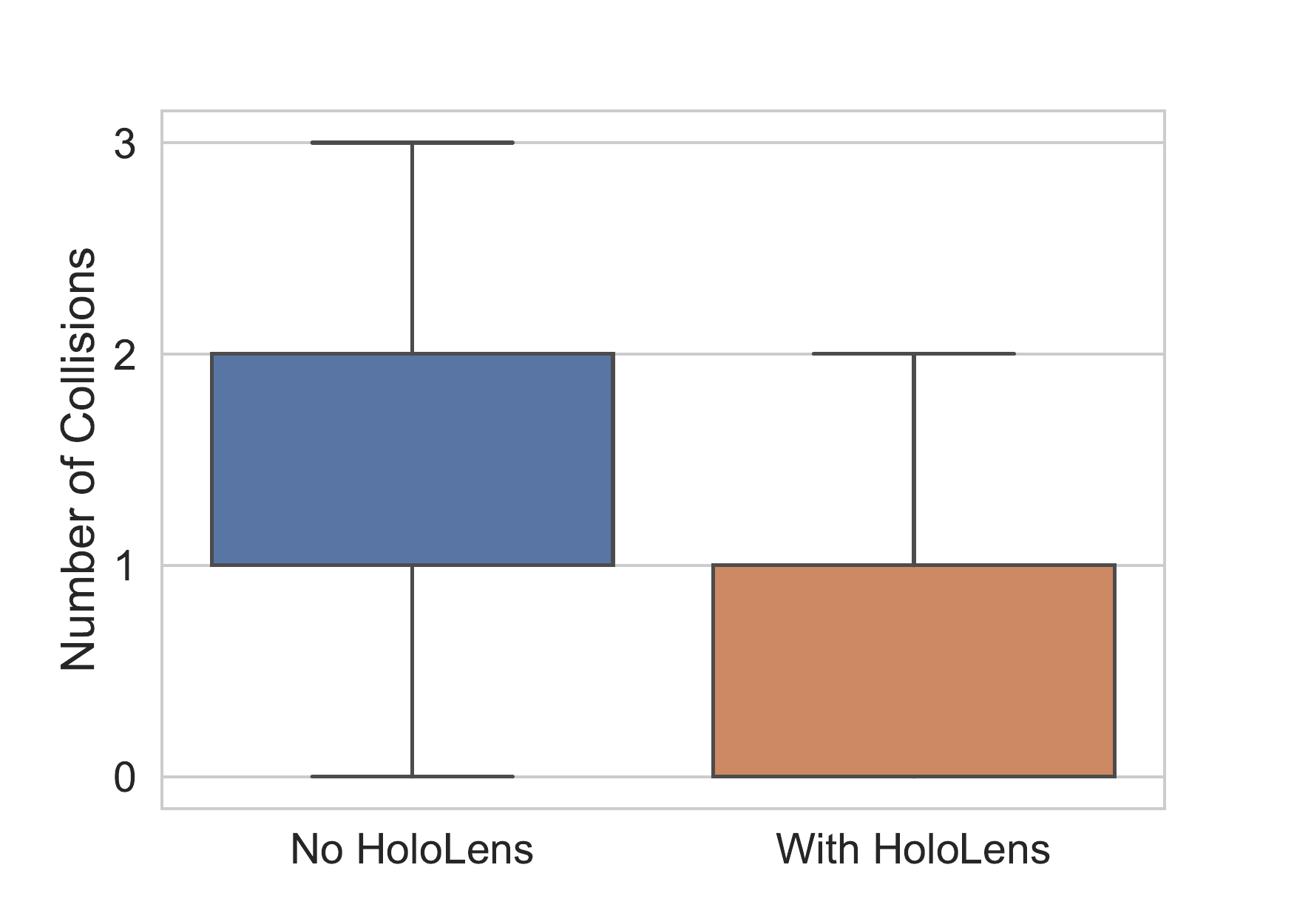}  
  \caption{Number of collisions per trial}
  \label{fig:collisions}
\end{subfigure}
\begin{subfigure}{.317\textwidth}
  \centering
  \includegraphics[width=\linewidth, trim={0.5cm 0.8cm 1.7cm 1.5cm},clip]{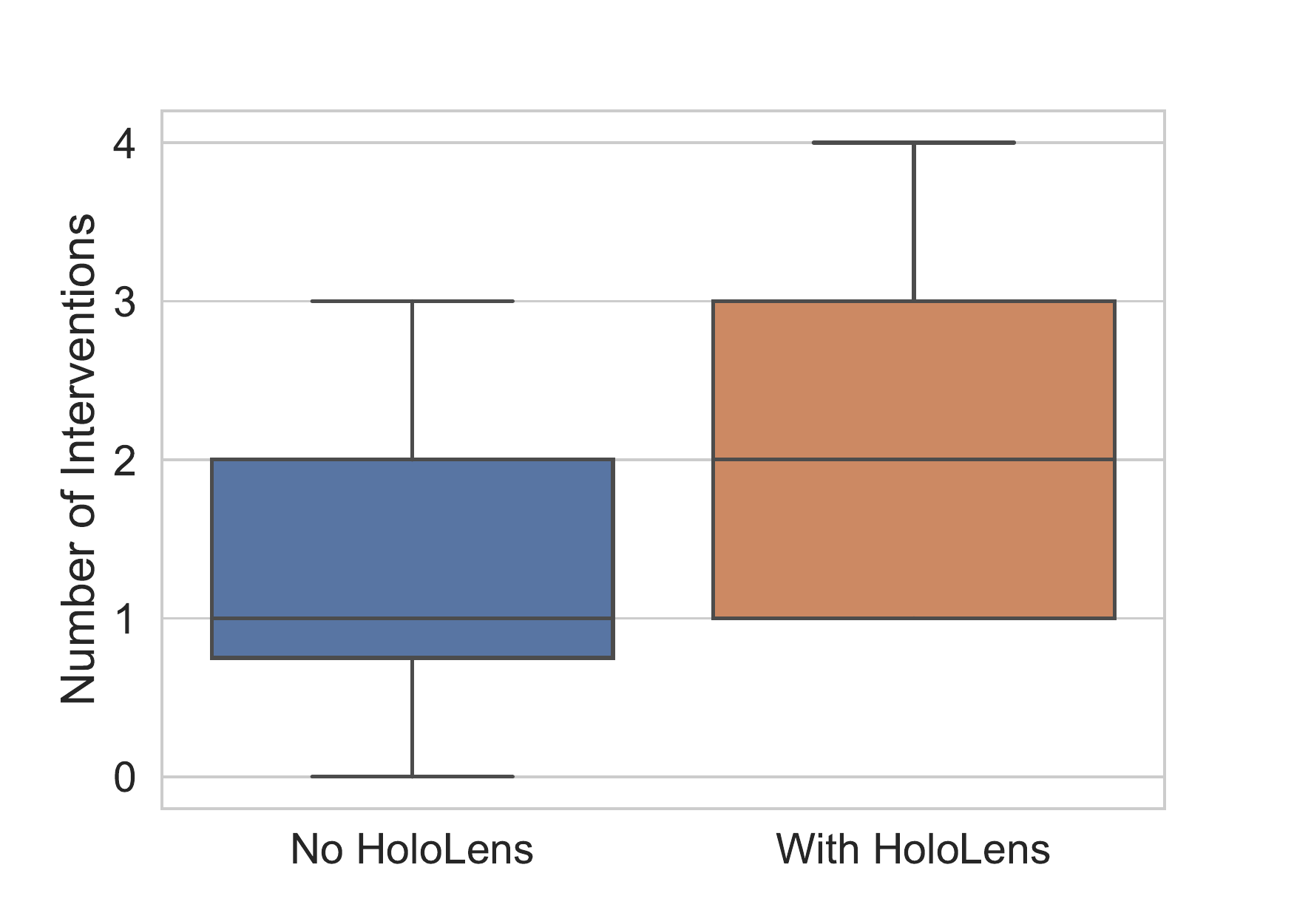}  
  \caption{Number of interventions per trial}
  \label{fig:interventions}
\end{subfigure}
\begin{subfigure}{.325\textwidth}
  \centering
  \includegraphics[width=\linewidth, trim={0cm 0.8cm 1.78cm 1.5cm},clip]{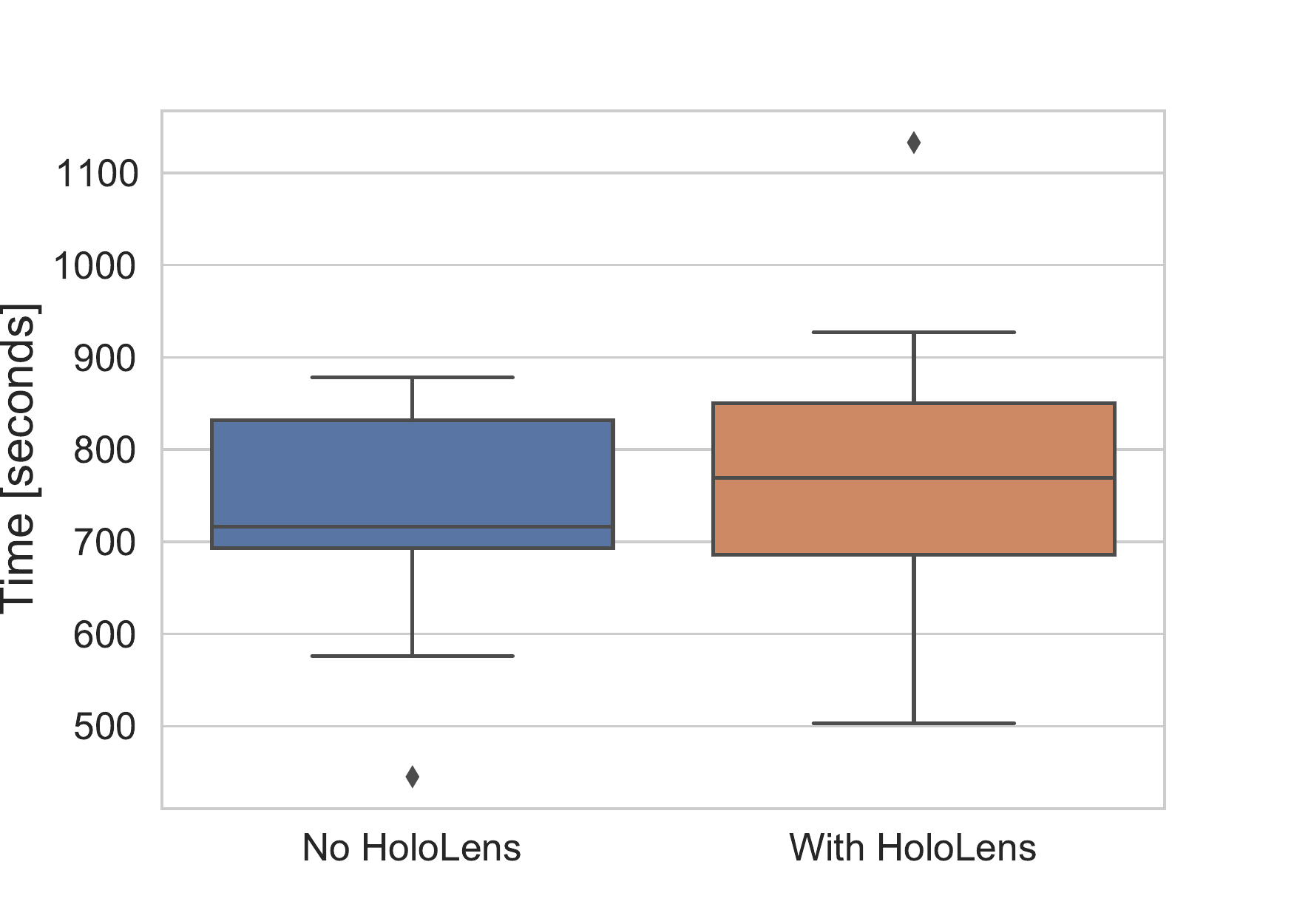}  
  \caption{Time required to complete the task}
  \label{fig:total_time}
\end{subfigure}
\caption{Results of the user study in the two experimental conditions. (a) and (b) show improvements in the trials carried out with the MR equipment, i.e., the HoloLens: at least half the population managed to complete the collaboration without experiencing collisions and at the same time participants felt more inclined to intervene and assist the robot; conversely, (c) shows no significant improvement in the adoption of the MR feedback in terms of task speed execution.}
\label{fig:exp_results}
\end{figure*}

\subsection{User Study} 
\label{experiments}

A pilot user study has been conducted with $S = 12$ subjects ($10$ males and $2$ females), aged between 24-38 (\textit{Avg} = $27.08$, \textit{StdDev} = $3.62$), with little to no prior experience with MR-HMDs. 
Each participant was requested to complete the collaborative assembly task with Baxter in two distinct conditions, namely:
\begin{enumerate}
\item[\textit{C1}] wearing the HMD, and with the \textit{am} medium activated, i.e., involving the whole communicative act $\hat{C}^*$;
\item[\textit{C2}] without wearing the HMD, only with the \textit{m} medium active, therefore with the communicative act $\hat{C}$ only.
\end{enumerate} 
In order to avoid introducing unwanted biases, each participant performed the two trials in reverse order with respect to the predecessor. 

Before beginning the trial in condition \textit{C1}, participants were provided with a brief tutorial to understand how to navigate within the HoloLens menus. 
Once settled and comfortable, they were asked to stand in front of the robot, open the Unity application from within the main menu, and then proceed to look at the 2D marker attached to the Baxter's front, thus enabling the spawning of the various holograms. 
They were then asked to align the various wooden pieces, tools and components involved in the assembly task with their virtual counterparts visible in the holographic scene. 
After that, they were able to start their trial by pressing the joystick's lever. 
During the trial performed in condition \textit{C2}, the various items and components were adjusted and aligned for each participant in advance.

During trials, in order to evaluate hypothesis \textit{H1.a}, the time required to perform each collaborative step, and the total time necessary to complete the assembly task were measured. 
Moreover, video recordings of the experiments were made, and later used to extract metrics useful to evaluate hypotheses \textit{H1} and \textit{H2}. 
In particular, within each trial we counted:
\begin{itemize}
\item For \textit{H1}, the number of unintentional collisions happening between the participant and Baxter, or between the robot and any objects involved in the assembly process;
\item For \textit{H2}, the number of times the participant proactively intervened to help the robot teammate not to fail an action, e.g., by adjusting the pose of an object to facilitate grasping.
\end{itemize}


\section{Results}
\label{sec:results}

The results obtained in the user study are summarized in Figure \ref{fig:exp_results}, where data related to collisions, intervention and time are presented.

The box plots in Figure \ref{fig:collisions} show the measured number of unintentional collisions during each trial. 
Due to the limited size of the shared workspace, collisions were more likely to happen during later stages of the assembly process, when the table was mostly cluttered by the half-built chair. 
Participants performing the experiment in condition \textit{C2} were more prone to be caught off-guard by the robot movements, thus resulting in an increased number of collisions. 
On the contrary, participants performing the trial in condition \textit{C1} were more aware of their robot teammate's presence.
Therefore, they knew better how to position themselves around the workspace while the interaction unfolded.
From the box plots in Figure \ref{fig:collisions}, one can notice that the data do not follow a normal distribution, therefore hypothesis \textit{H1} has been evaluated through a non-parametric test, namely the one-tailed Wilcoxon signed-rank test \cite{wilcoxon1992individual}. 
The test provided a statistic $W = 3.5$ and $p < 0.01$. 
This result was compared with the critical value $W_c$ extracted from the table in \cite{wilcoxon1947probability} by fixing the significance level $\alpha = 0.05$ and the number of participants $S$, thus yielding $W_c = 17$. 
Since $W < W_c$, the test allowed us to reject the null hypothesis and conclude that there is a statistical difference between the two experimental conditions.

Similarly, Figure \ref{fig:interventions} depicts the measured number of proactive interventions per trial. 
In this case, participants in condition \textit{C1} were able to understand the robot intentions in advance, and thus were more likely to help Baxter grasp an ill-positioned tool or a wooden piece by slightly adjusting its position on the table. 
Conversely, in condition \textit{C2}, users could not anticipate what the robot was about to do and, as such, felt less prone to intervene and help it. 
This resulted also in the robot possibly failing a grasping action, consequently damaging the overall quality of the interaction. 
As before, hypothesis \textit{H2} has been evaluated through the Wilcoxon test, which yielded a statistic $W = 9.5$ with $p< 0.04$. 
Comparing $W$ with $W_c$ enabled us to reject the null hypothesis and to state that the two experimental conditions are statistically different.

Figure \ref{fig:total_time}, however, shows that the amount of time required to complete the assembly task is not considerably affected by the adoption of the MR setup. 
This is also confirmed by the test, which does not yield significant results ($W = 27.0$, $p= 0.381$). 
For this reason, we could not reject the null hypothesis for \textit{H1.a}. 
This behavior could be explained as a result of the collaborative task's structure, our explanation being as follows.
On the one hand, robot actions are scripted, therefore the time required for their completion is unchanged with respect to the experimental condition while, on the other hand, each participant performed their assigned assembly steps in roughly the same amount of time over the course of the two trials. 
Therefore, although the number of collisions is indeed higher in condition \textit{C2}, the seconds lost by the robot to recover after every accidental bump do not represent a significant deterioration of the overall task execution time over the course of a $10$ or more minutes assembly process.


\section{Conclusions}
\label{sec:conclusions}

Communicating the robot intentions and movements to a human teammate in an intuitive way is a fundamental aspect to achieve a fluent and natural human-robot collaboration. 
In this paper, we postulate that conveying dynamic information about upcoming robot motions may be a possible step towards such a goal. 
Therefore, we presented a modular software architecture enabling a human teammate to visualize in advance robot's future motions through MR equipment, i.e., by exploiting the power of modern HMDs. A user study has been carried out to assess whether a complex communicative act from the robot to the human teammate, which results from the combination of \textit{normal} robot motions and \textit{anticipatory} robot motions conveyed via the holographic medium, could effectively improve the collaboration process.
The results and statistical tests presented in the paper allowed us to conclude that such form of communication made participants more inclined to assist their robot teammate, and enabled them to achieve a collaboration less subject to collisions. 
However, results also showed no significant improvement in the task execution speed while using MR equipment, therefore we could not verify an initial hypothesis related to a possible improvement of speed in the collaboration process. 
We conjecture that the MR medium could have greater influence on the task execution speed when employed in more variable collaborative scenarios where the robot does not follow a scripted action plan.

In light of the restricted number of participants who where involved in this study, we plan to perform additional tests to evaluate whether our findings can be generalized to a broader population, as well as to carry out analyses on the user experience of the MR interface by means of questionnaires and other evaluation metrics, e.g., wearable sensors \cite{bethel2010review}. Although the MR-HRC architecture is general and only requires the robot platform to be ROS-enabled, future works will also focus on testing its generalization capabilities with multiple robot models and on integrating it in a flexible collaborative scenario where human and robot concurrently work together without a predetermined sequence of actions.




\bibliographystyle{IEEEtran}
\bibliography{bibliography}

\end{document}